\documentclass[10pt,twocolumn,letterpaper]{article}

\usepackage{iccv}
\usepackage{times}
\usepackage{epsfig}
\usepackage{graphicx}
\usepackage{amsmath}
\usepackage{amssymb}

\usepackage{subcaption}
\usepackage{booktabs}
\usepackage[british,american]{babel}
\usepackage{amsfonts}
\usepackage{color}
\usepackage{stfloats}
\usepackage{balance}
\usepackage[pagebackref=true,breaklinks=true,letterpaper=true,colorlinks,bookmarks=false]{hyperref}
\newcommand{\emoji}[1]{\includegraphics[height=1em]{./#1.png}}
\newcommand{\figref}[1]{\mbox{Fig.~\ref{#1}}}
\newcommand{\tblref}[1]{\mbox{Table~\ref{#1}}}
\newcommand{\secref}[1]{\mbox{Sec.~\ref{#1}}}
\renewcommand{\eqref}[1]{\mbox{Eq.~\ref{#1}}}
\let\oldparagraph\paragraph
\renewcommand{\paragraph}[1]{\vspace{-0.4cm} \oldparagraph{#1}}
\newcommand*{\email}[1]{\tt\small{#1}}

\newif\ifedit

\iccvfinalcopy 

\def\iccvPaperID{}

\ificcvfinal\fi
\begin{document}

\title{Smile, Be Happy :) Emoji Embedding for Visual Sentiment Analysis}

\author{
\begin{tabular}{cccc}
Ziad Al-Halah & Andrew Aitken & Wenzhe Shi & Jose Caballero \\
\email{ziadlhlh@gmail.com} & \email{aaitken@twitter.com} & \email{wshi@twitter.com} & \email{jcaballero@twitter.com} \\
\multicolumn{4}{c}{Twitter}\\
\end{tabular}\\
}

\maketitle

\begin{abstract}
Due to the lack of large-scale datasets, the prevailing approach in visual sentiment analysis is to leverage models trained for object classification in large datasets like ImageNet.
However, objects are sentiment neutral which hinders the expected gain of transfer learning for such tasks.
In this work, we propose to overcome this problem by learning a novel sentiment-aligned image embedding that is better suited for subsequent visual sentiment analysis.
Our embedding leverages the intricate relation between emojis and images in large-scale and readily available data from social media.
Emojis are language-agnostic, consistent, and carry a clear sentiment signal which make them an excellent proxy to learn a sentiment aligned embedding.
Hence, we construct a novel dataset of $4$ million images collected from Twitter with their associated emojis.
We train a deep neural model for image embedding using emoji prediction task as a proxy.
Our evaluation demonstrates that the proposed embedding outperforms the popular object-based counterpart consistently across several sentiment analysis benchmarks.
Furthermore, without bell and whistles, our compact, effective and simple embedding outperforms the more elaborate and customized state-of-the-art deep models on these public benchmarks.
Additionally, we introduce a novel emoji representation based on their visual emotional response which supports a deeper understanding of the emoji modality and their usage on social media.
Project page: \url{https://www.cs.utexas.edu/~ziad/emoji_visual_sentiment.html}.

\end{abstract}

\section{Introduction}
Analyzing people's emotions, opinions, and attitudes towards a specific entity, an event or a product is referred to as sentiment analysis~\cite{medhat2014sentiment,liu2012survey}.
Sentiment can be reduced to \textit{positive}, \textit{neutral}, and \textit{negative}, or can be extended to a richer description of fine-grained emotions, such as \textit{happiness}, \textit{sadness}, or \textit{fear}.
Summarizing and understanding sentiment has important applications in various fields like interpretation of customer reviews, advertising, politics, and social studies.
Thus, automated sentiment analysis is an active subject of research to devise methods and tools to enable such applications~\cite{kim2018building,peng2015mixed,alameda2016recognizing}.

\begin{figure}[!t]
\vspace{-0.3cm}
	\centering
    \includegraphics[width=0.8\linewidth]{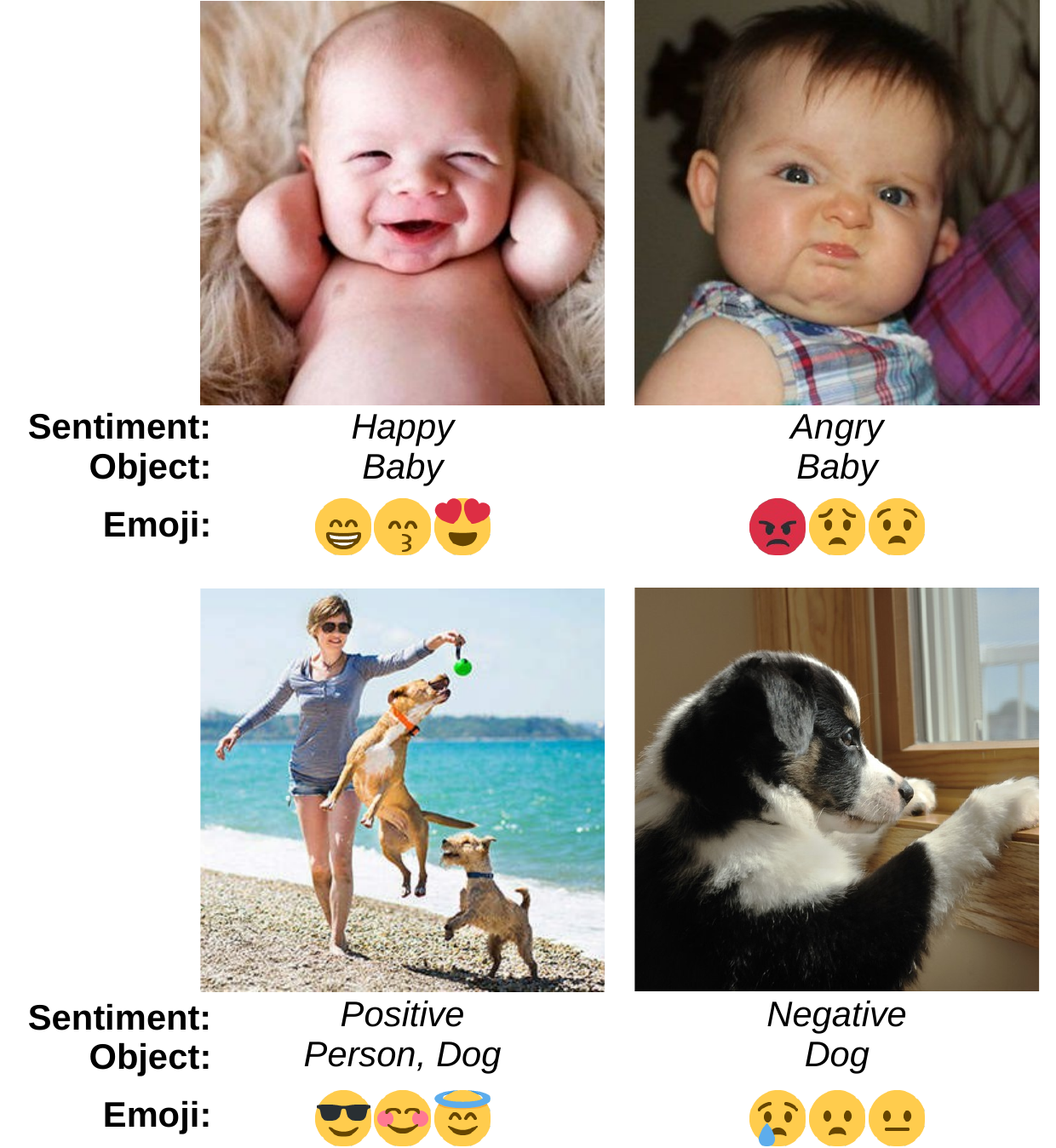}    
	\caption{Images with similar objects may show different sentiments. 
	Unlike the object neutral representation, emoji embedding is well aligned with the sentiment label space. 
	Hence, it is expected to generalize well in transfer learning settings for visual sentiment and emotion analysis.}
\label{fig:intro_smiley}
\end{figure}
 
Driven by the availability of large-scale annotated datasets~\cite{glorot2011domain,tang2014coooolll} along with modern deep learning models, language sentiment analysis witnessed great improvements over the last few years~\cite{mohammad2018semeval}.
However, \textit{visual} sentiment analysis still lags behind.
This is mainly due to the lack of large-scale image datasets with sentiment labels.
Current datasets (\eg,~\cite{you2015robust,ng2015deep,alameda2016recognizing,lang2005international,machajdik2010affective}) are scarce and too small to appropriately train deep neural networks, which are prone to overfitting the small training data.

To overcome the previous problem, the dominant approach currently is to employ cross-domain transfer learning methods.
This is achieved by \textit{pretraining} a deep neural network on a large-scale dataset for object classification, such as ImageNet~\cite{russakovsky2015imagenet}, and then \textit{fine-tuning} the network for sentiment classification on the small target dataset.
This approach is unanimously adopted by recent visual sentiment models and has led to improved results, \eg~\cite{you2015robust,campos2017pixels,ng2015deep}.
Nonetheless, object categories and sentiment labels are not aligned and rather orthogonal.
Object labels are sentiment neutral; \ie objects of the same category can exhibit various emotions (\figref{fig:intro_smiley}).
Hence, the domain gap between object recognition and sentiment analysis is significant.
Pretraining a model with an object-focused embedding may not be the most useful representation for subsequent transfer learning for sentiment or emotion classification.

Given that collecting data for the target task is impractical, is there an alternative representation which 1) is better aligned with sentiments and 2) can be learned efficiently with minimum overhead?
Emojis, with the advent of social media, became a prevailing medium to emphasize emotions in our communications such as happiness \emoji{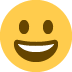}, anger \emoji{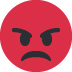}, or fear \emoji{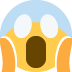}.
Not only do emojis carry a clear sentiment signal by themselves (see \figref{fig:intro_smiley}), they also act as sentiment magnifiers or modifiers of surrounding text~\cite{novak2015sentiment}.
Additionally, due to their prominent use in social media like Facebook, Twitter and Instagram, one can relatively easily tap into large amounts of readily available data without the need for any manual labeling.
All these factors turn an emoji-based representation into an attractive candidate for our target task of visual sentiment analysis.
In fact, emojis have been successfully leveraged for \textit{language} sentiment analysis recently~\cite{guthier2017language,felbo2017using,rathan2018consumer}.

However, the interaction among emojis and the corresponding images in social media remains elusive.
Is there a strong correlation between an emoji and a visual signal? And if so, do emojis capture the visual sentiment exhibited in images?
The answer to these questions is not straightforward.
Social media data is known to be noisy~\cite{baldwin2013noisy}, and the use of emojis is influenced by the user's cultural background~\cite{barbieri2016cosmopolitan,ljubevsic2016global} and major temporal events~\cite{santhanam2018stand}.
These hurdles represent important challenges to learning an effective emoji representation that can generalize well across domains.
In this paper, we present the \textit{first} work to address the previous questions with a thorough analysis of emojis and their visual sentiment connotation.

To that end, we leverage weakly labeled data collected from social media (\eg Twitter) to build a large-scale dataset of $4$ million images and their corresponding emoji annotation.
Through extensive experiments, we demonstrate that an emoji based representation can be effectively learned from such noisy data.
Moreover, using off-the-shelf deep neural models and without bells and whistles, we show that our emoji embedding exhibits remarkable generalization properties across domains and outperforms state-of-the-art in visual sentiment and fine-grained emotion recognition.
Additionally, we introduce a new perspective on emoji interpretation using their visual emotional signature and their perceived similarity in the visual emotion space.

\section{Related Work}
\vspace{0.2cm}

\paragraph{Visual sentiment analysis}
While sentiment analysis from text has been extensively studied, extracting sentiment from visual data has proven to be more challenging, primarily due to the lack of large-scale datasets suited for advanced models like deep neural networks.
Most available datasets are small and contains only hundreds (\eg~\cite{mikels2005emotional,you2015robust}) or a few thousands (\eg~\cite{you2016building}) samples.
Hence, many visual sentiment methods rely on hand-crafted features (\eg color histograms, SIFT) to train simple models with few parameters in order to avoid the risk of overfitting the training data~\cite{machajdik2010affective,lu2012shape,zhao2014exploring}.
However, it is hard for such low-level features to effectively capture the higher level concept of sentiment.
One way to overcome the previous problem is by learning an intermediate representation from external data that helps bridging the gap between low-level features and sentiment.
For example, this can be achieved by learning an intermediate concept classifier for Adjective Noun Pairs (ANP) as in the SentiBank model~\cite{borth2013large}.
However, the most common approach is to take advantage of powerful models, \ie deep neural networks, in a transfer learning setting~\cite{you2015robust,campos2017pixels,yang2018visual}.
In this case, the neural network model is initially trained on a large-scale dataset for object classification~\cite{russakovsky2015imagenet}.
Afterwards, the model is fine-tuned on the target task for sentiment prediction.

However, while ANP- and object-based embedding lead to improved performance, both are still not ideal for sentiment analysis.
It is not clear how to select a good ANP vocabulary that can generalize well to various tasks requiring the inference of emotions from images.
Additionally, object-based models are not suited for capturing sentiment since they are trained for \textit{sentiment neutral} object classification.
In this work, we propose to learn an emoji-based embedding for cross-domain sentiment and emotion analysis.
Unlike objects and ANPs, emojis carry a strong sentiment signal which leads to a compact and powerful representation outperforming the previous methods as demonstrated by our evaluation.

\paragraph{Emojis}
Due to the increasing popularity of emojis, there is great interest in analyzing and studying their usage, \eg~\cite{kelly2015characterising,lebduska2014emoji,miller2017understanding,ljubevsic2016global}.
Most of this work is carried from a natural language processing (NLP) point of view, \eg~\cite{barbieri2016does,eisner2016emoji2vec}.
More relevant to our work is the analysis of emojis and sentiment.
Emojis can be shown to act as a strong sentiment signal that generalizes well when analyzed from a NLP perspective~\cite{read2005using,go2009twitter,novak2015sentiment,felbo2017using,rathan2018consumer,guthier2017language}.
However, whether the same can be said for a visual sentiment perspective is still to be determined.
Recently, few studies attempted to learn the correlations between the emoji and image modalities.
In \cite{el2017face2emoji}, a model is developed to predict the proper emoji matching a facial expression input.
On the other hand, \cite{cappallo2018new} propose to handle emojis as new modality and introduce a model to predict visual or textual concepts by using emojis correlations, \eg learn a ship classifier by leveraging the ship emoji~\emoji{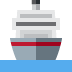}.
In contrast to previous work, and to the best of our knowledge, this work is the first to propose emoji embedding for cross-domain visual sentiment analysis and provide an in depth analysis of their visual sentiment and emotional interpretation.

\section{Emoji for Visual Sentiment Analysis}
\label{sec:approach}

We aim in this work to learn an efficient and low-dimensional embedding of images in the emoji space.
This embedding is well aligned with and encodes the visual sentiment exhibited in an image.
Moreover, it can be learned efficiently from large-scale and weakly labeled data.
To that end, we introduce a large-scale benchmark for visual emoji prediction (\secref{sec:dataset}) along with  deep neural model for efficient emoji embedding and transfer learning (\secref{sec:model}).

\subsection{Visual Smiley Dataset}
\label{sec:dataset}

In this section, we describe our method for data collection from social media, including a) the selection of emoji categories; b) the analysis of the sample distribution; and c) a temporal sampling strategy that suits our learning task\footnote{The visual smiley dataset collected and used as part of this work is available at: \href{https://twitter.app.box.com/v/visual-smiley-dataset}{twitter.app.box.com/v/visual-smiley-dataset}}.

\paragraph{Categories}
The emoji list has grown from $76$ entries in $1995$ to $3019$ in the latest \textit{Emoji v12.0} in $2019$~\cite{emojiv12}.
Many of these emojis represent objects categories (\eg \emoji{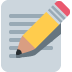}\emoji{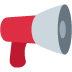}\emoji{ship}), abstract concepts (\eg \emoji{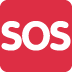}\emoji{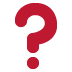}\emoji{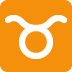}) or animals and plants (\eg \emoji{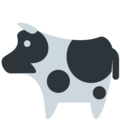}\emoji{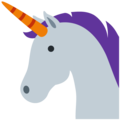}\emoji{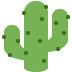}).
These types of emojis are either sentiment neutral or have weak correlation with sentiment that usually arise from users cultural background or personal preferences, \eg towards certain animal classes.
Since our goal is to have an emoji-based representation for sentiment analysis these types are excluded from our selection.
As our target categories, we chose a subset of $92$ popular emojis which commonly referred to as \textit{Smileys} (\eg \emoji{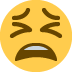}\emoji{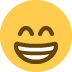}\emoji{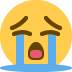}).
These smileys show a clear sentiment or emotional signal which make them adequate for our cross domain sentiment analysis.
Moreover, they are among the most frequently used emojis in social media which further facilitates data collection and aids the learning process.

\paragraph{Sample Distribution}
Social media such as Instagram, Flickr and Twitter represent a rich source for large-scale emoji data.
It is estimated that more than $700$ million emojis are sent daily over Facebook while half the posts in Instagram contains emojis~\cite{emoji_stats}.
Here, we select our samples from Twitter such that we target only tweets that contain emojis and are associated with at least one image.
Furthermore, to increase the relevance between the emojis and the associated image in the samples we constrain the selected tweets to those that do not contain urls, hashtags nor user mentions.
This is motivated by the observation that these elements usually represent important context cues to understand the use of the selected emoji that goes beyond the associated visual data.
We additionally ignore tweets that are quotes or replies to other tweets to reduce redundancy.

\begin{figure}[!t]
\centering
\begin{subfigure}[!b]{0.95\linewidth}
    \centering
    \includegraphics[width=\linewidth]{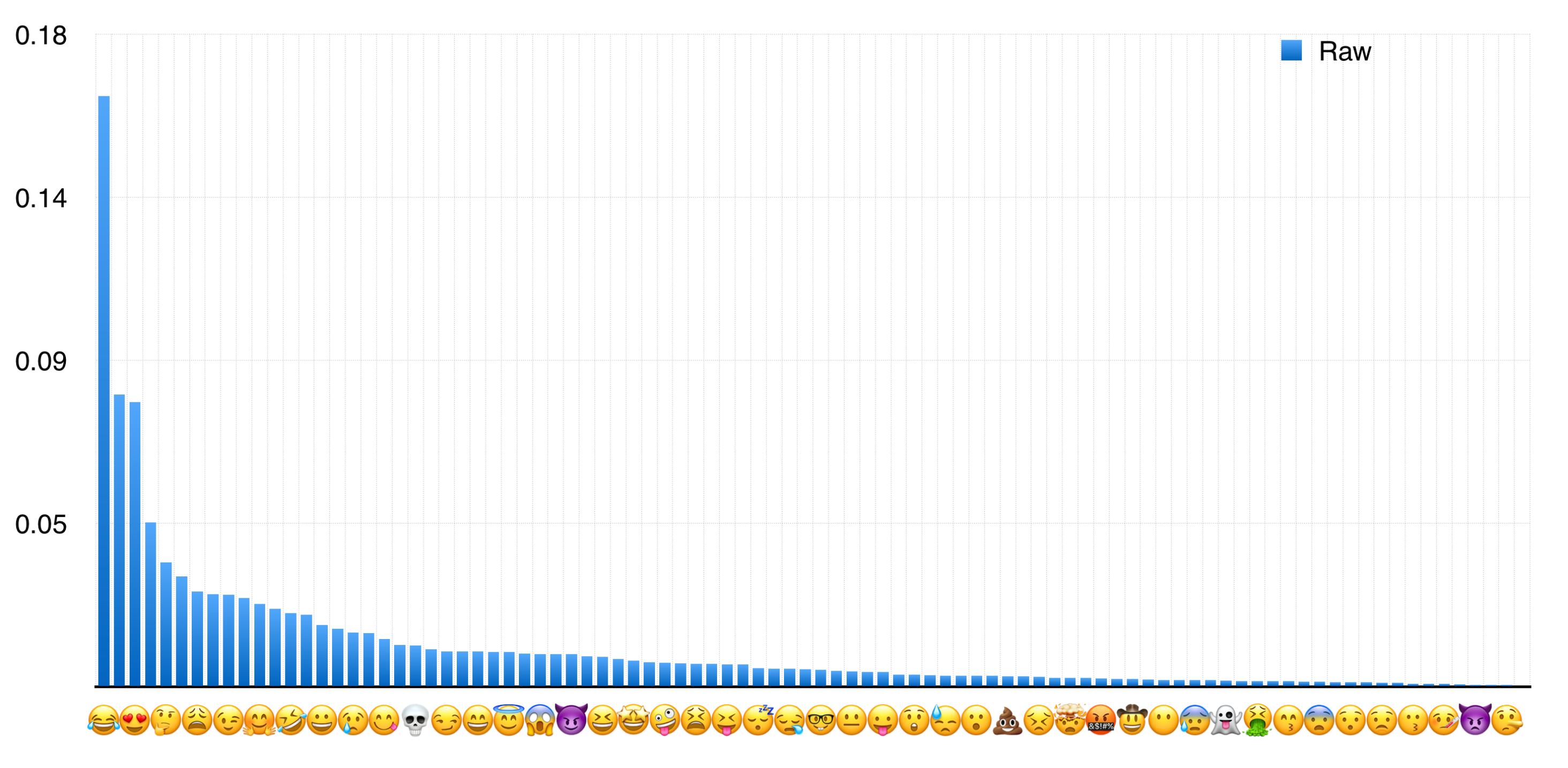}    
    \caption{Raw data distribution}
\end{subfigure}\\
\begin{subfigure}[!b]{0.95\linewidth}
    \centering
    \includegraphics[width=\linewidth]{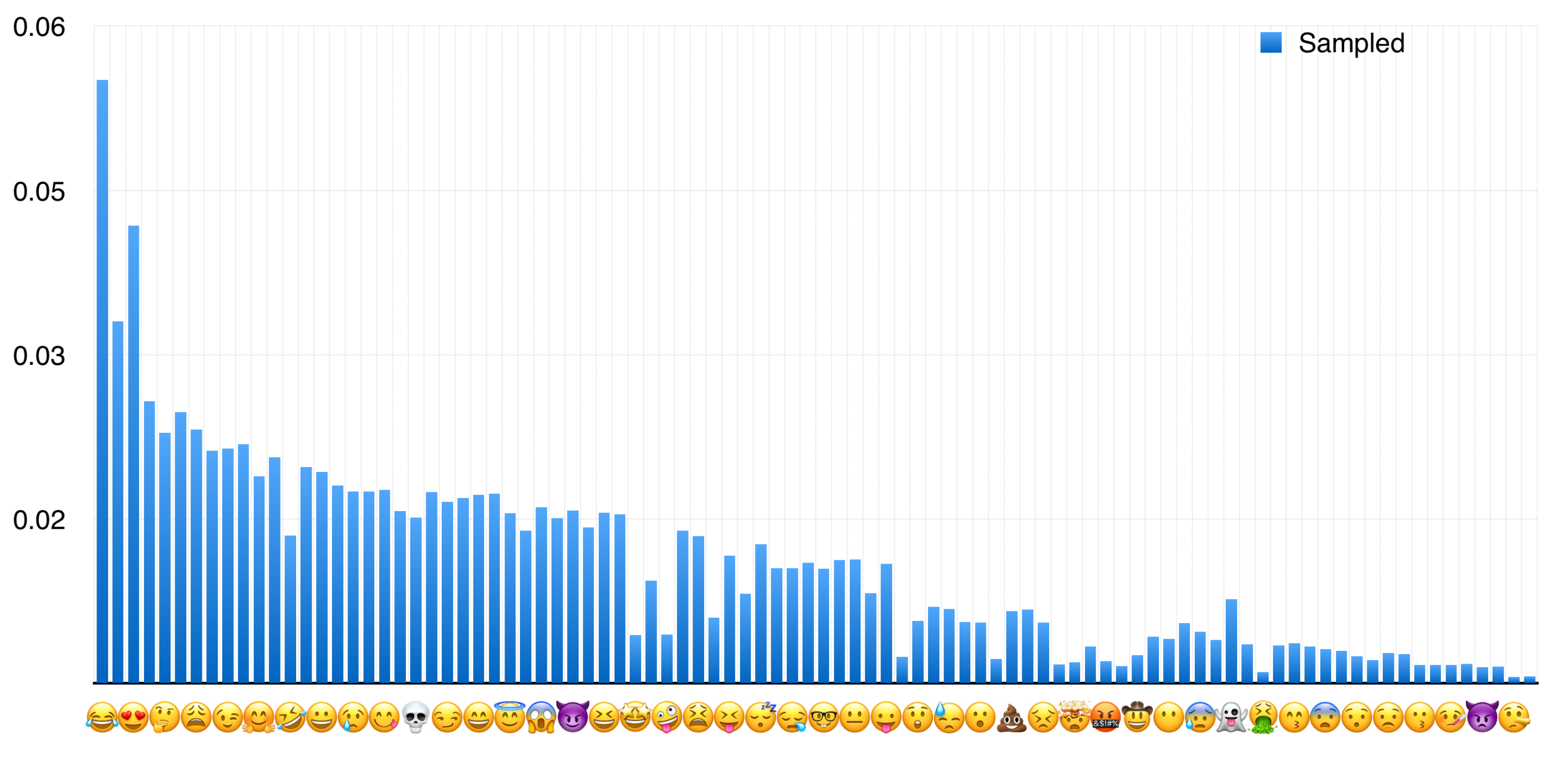}    
    \caption{Temporally balanced data distribution}
\end{subfigure}
\caption{Emoji frequency in (a) a raw sample of data and (b) the temporal balanced sampled dataset. Dataset (b) is used in this study.}
\label{fig:emoji_dist}
\vspace{-0.3cm}
\end{figure}
 Given the previous criteria, we retrieve a collection of $2.8$ million Tweets from the first six months of $2018$.
\figref{fig:emoji_dist}a shows the label distribution of the data.
We see that this data has a long-tail distribution and is heavily biased towards a few categories, with the top $5$ most frequent emojis (\ie \emoji{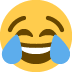}\emoji{043}\emoji{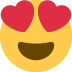}\emoji{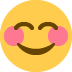}\emoji{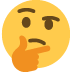}) representing around $40\%$ of the retrieved samples.
This poses a great challenge for most standard machine learning methods as an imbalanced training dataset may lead a training process to trivially predict the most frequent labels instead of learning a more meaningful representation.
Additionally, we notice that when collecting the data from a relatively short time period the content of samples tends to be heavily biased towards a few major temporal events (\eg USA presidential elections or World Cup).
This in turn reduces the variability of the images and hence the ability of the model to generalize well across domains.

\paragraph{Temporal Sampling}
To overcome content homogeneity, we propose to retrieve the samples from a relatively large time period while uniformly sampling the tweets from smaller temporal windows.
Specifically, we collect tweets from January $1^{st}$ $2016$ till July $31^{st}$ $2018$.
We split the time range to sequential time windows of $30$ days.
Furthermore, to alleviate label imbalance we randomly select a maximum of $4000$ tweets for each emoji category within each window.
We additionally allow valid samples to have a maximum number of $5$ emojis, meaning that certain samples will contain multiple labels.
In total, this methodology led to about $4$ million images with $5.2$ million emoji labels.
\figref{fig:emoji_dist}b shows the label distribution of the sampled dataset.
We see that compared to the raw data distribution, our dataset is more balanced across the various categories.
Nonetheless, some emojis still occur relatively more often than others due to the multi-label nature of the data and the innate inter-emoji correlations.

To get a better notion of the correlation between labels, we construct the normalized correlation matrix of all emojis in the collected data\footnote{See appendix for the full correlation matrix}.
As expected, by analyzing the correlation matrix we see that the two most frequent emojis \emoji{006} and \emoji{015} co-occur with most of the categories.
Additionally, the correlation matrix reveals some semantically related groups like [\emoji{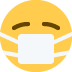}\emoji{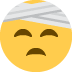}\emoji{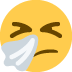}\emoji{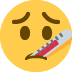}\emoji{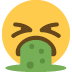}] and [\emoji{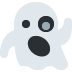}\emoji{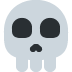}\emoji{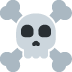}].

\subsection{Smiley Embedding Network}
\label{sec:model}

\begin{figure}[!t]
	\centering
    \includegraphics[width=\linewidth]{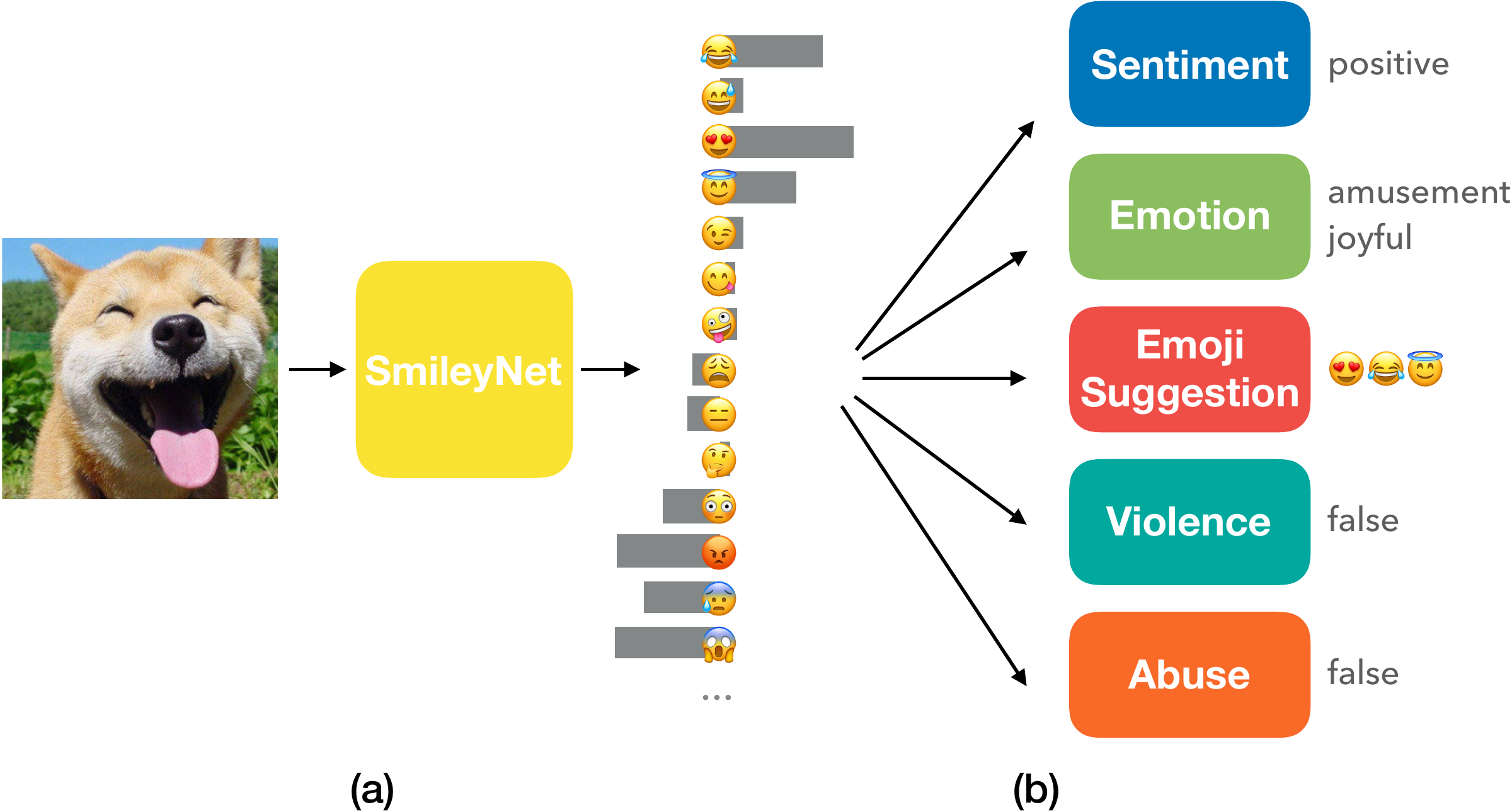}
	\caption{Our model (SmileyNet) (a) learns to embed images in the low-dimensional emoji space from large-scale and noisy data collected from social media. This embedding can subsequently be leveraged via transfer learning (b) for many target tasks in which deriving emotions from visual data is needed, such as sentiment and emotion analysis.}
\label{fig:model}
\vspace{-0.4cm}
\end{figure}
 Given the large-scale nature of the collected dataset, it is possible to leverage deep neural network architectures for effective learning of the emoji embedding with reduced risks of data overfitting.
Formally, our goal is to learn an embedding function $f(\cdot)$ that maps an image $\mathbf{x}\in \mathcal{X}^{d_x}$ to an embedding in the emoji space $\mathbf{e} \in \mathcal{E}^{d_e}$, \ie $f: \mathcal{X}^{d_x}\rightarrow\mathcal{E}^{d_e}$.
Such that $d_x$ and $d_e$ are the dimensionality of the image and emoji spaces respectively.
An efficient option to realize $f(\cdot)$ is through the proxy task of explicit emoji prediction (\figref{fig:model}a).
This has two main advantages compared to other options like metric learning in the emoji space.
Firstly, it is more computationally efficient compared to Siamese and Triplet networks that are usually employed for metric learning.
Hence, it scales easily to large datasets while using less resources.
Secondly, the learned embedding through the emoji prediction task is interpretable since each dimension in $\mathbf{e}$ corresponds to one of the emoji categories, \ie $d_e = \mathit{C}$ where $\mathit{C}$ is the number of emoji categories.
This enables subsequent analysis of the embedding, better understanding of model properties, and a novel zero-shot visual sentiment learning task as we will see in \secref{sec:eval}. 

To that end, we train an emoji prediction model $h(\cdot)$ such that:
$h(\mathbf{x}) = \sigma(f(\mathbf{x}))$,
where $\sigma$ is the sigmoid activation function since our task is a multi-label classification problem.
Then $h(\cdot)$ can be optimized using the binary cross entropy loss:
\begin{equation}
	\mathcal{L}(\mathbf{x}_i, \mathbf{y}_i)=-\sum_{c=1}^\mathit{C} \mathbf{y}_{i,c}~log(h(\mathbf{x}_i)_c),
\end{equation}
where $\mathbf{y}_{i,c}$ is the binary label for the emoji of class $c$, and $h(\mathbf{x}_i)_c$ is the probability of the model predicting class $c$ for image $\mathbf{x}_i$.

\paragraph{Transfer learning}
Once $f(\cdot)$ is trained, we can easily adapt our model across domains for a target task $g(\cdot)$ such as sentiment or emotion prediction (\figref{fig:model}b).
This is achieved through $t(\cdot)$ that maps the emoji embedding to the target label space $\mathcal{T}$, such that
$g=t\circ f: \mathcal{X}\rightarrow \mathcal{E} \rightarrow \mathcal{T}$.
$t(\cdot)$ is realized using a multilayer perceptron and $g(\cdot)$ can then be learned using the small training data of the target task.

\section{Evaluation}
\label{sec:eval}
We evaluate our embedding model (\textit{SmileyNet}) for three main tasks: 1) emoji prediction which is used as a proxy to train our embedding model; and the transfer learning tasks of 2) visual sentiment analysis and 3) fine-grained emotion classification.
Furthermore, 4) we introduce and analyze a novel representation for emojis that captures their unique properties in the visual sentiment space.

\subsection{Emoji Prediction}
\vspace{0.4cm}
\paragraph{Implementation}
Given our visual smiley dataset, we select $45$ thousand images for validation and $91$ thousand for testing.
Samples in the validation and testing splits are balanced such that each category has around $500$ and $1000$ samples, respectively.
We use the remaining data to train our SmileyNet model.
We adopt a residual neural network with $50$ layers ResNet50~\cite{he2016deep} as the base architecture for SmileyNet.
The model parameters are estimated using Adam~\cite{Kingma2015} for stochastic gradient descent optimization with an initial learning rate of $1e-4$.
Furthermore, we leverage data augmentation during training by randomly selecting an image crop of size $224\times 224$ pixels with random horizontal flipping and scaling.
The model is trained for $320,000$ iterations with a batch size of $128$ images.

\paragraph{Evaluation metric}
Since emoji prediction is a multi-label task, we adopt a variant of the Top-$k$ accuracy that accounts for the number of correct emojis in the top $k$ predictions out of the set of ground truth emoji of each sample.
Formally:
\begin{equation}
    \text{mTop-}k_i(p_i, y_i) = \frac{|\mathrm{ind_{k}}(p_i)\cap\mathrm{ind}(y_i=1)|}{\mathrm{min}(k, |\mathrm{ind}(y_i=1)|)},
\end{equation}
where $p_i=p(y|x_i)$ is the model prediction given image $x_i$, $\mathrm{ind_{k}}(p_i)$ are the indexes of the top $k$ predictions, and $\mathrm{ind}(y_i=1)$ are the indexes of the ground truth labels.
Notice that here $p_i \in \mathbb{R}^\mathit{C}$ and $y_i \in \mathbb{R}^\mathit{C}$ are vectors in which $p_{i, c}$ and $y_{i, c}$ are individual entries.
The final $\text{mTop-}k$ is the average over all $N$ samples in the test split:
\begin{equation}
    \text{mTop-}k=\frac{1}{N}\sum_i^N \text{mTop-}k_i(p_i,y_i).
\end{equation}
We also report the average area under curve (AUC) of the receiver operating characteristic (ROC) of all categories.

\begin{table}
\center
\scalebox{0.80}{
\begin{tabular}{ l c c c c}
\toprule
Model 						& 	mTop-1 	& 	mTop-3 	& 	mTop-5 	& 	AUC \\
\midrule
Random performance			& 	1.7		&	3.3		&	5.4		&	50.0   \\
SmileyNet (Raw-Dist.)		& 	9.5 	& 	11.6 	& 	16.3 	& 	67.6   \\
SmileyNet (Temp-Sampling)	& 	\textbf{11.5} 	& 	\textbf{14.4} 	& 	\textbf{19.5} 	& 	\textbf{69.8}   \\
\bottomrule
\end{tabular}
}
\caption{Emoji prediction performance of our SmileyNet on the proposed Visual Smiley Datasets.}
\label{tbl:emoji_pred}
\vspace{-0.3cm}
\end{table}

 \paragraph{Results}
Along with the full model, we test two variants: 1) a random baseline and b) our SmileyNet trained with the raw emoji distribution (Raw-Dist.) without the proposed temporal sampling (\secref{sec:dataset}).
\tblref{tbl:emoji_pred} shows the performance of these models in emoji prediction on the testing split.
We notice that even with a noisy data source as social media, our model is able to predict emojis from images significantly better than a random baseline.
Furthermore, our temporal sampling method leads to higher performance, \ie better learned embedding, compared to a model learned with the raw and biased data distribution.
In general, we see that the accuracy is relatively low.
This can be attributed partly to the expected amount of noise in data annotations since it is collected automatically without any human intervention; and also to the strict evaluation metric adopted in this task which tend to underestimate the model performance.
For example, a prediction of \emoji{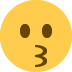} by our model for an image labeled with \emoji{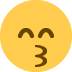} is considered wrong.
Additionally the model needs to predict all annotated emojis for an image to get a full score on it.
Nonetheless, our subsequent qualitative and transfer learning evaluation confirms that our SmileyNet in fact learns a compelling visual embedding with high performance.

\begin{figure}[!t]
	\centering
    \includegraphics[width=0.8\linewidth]{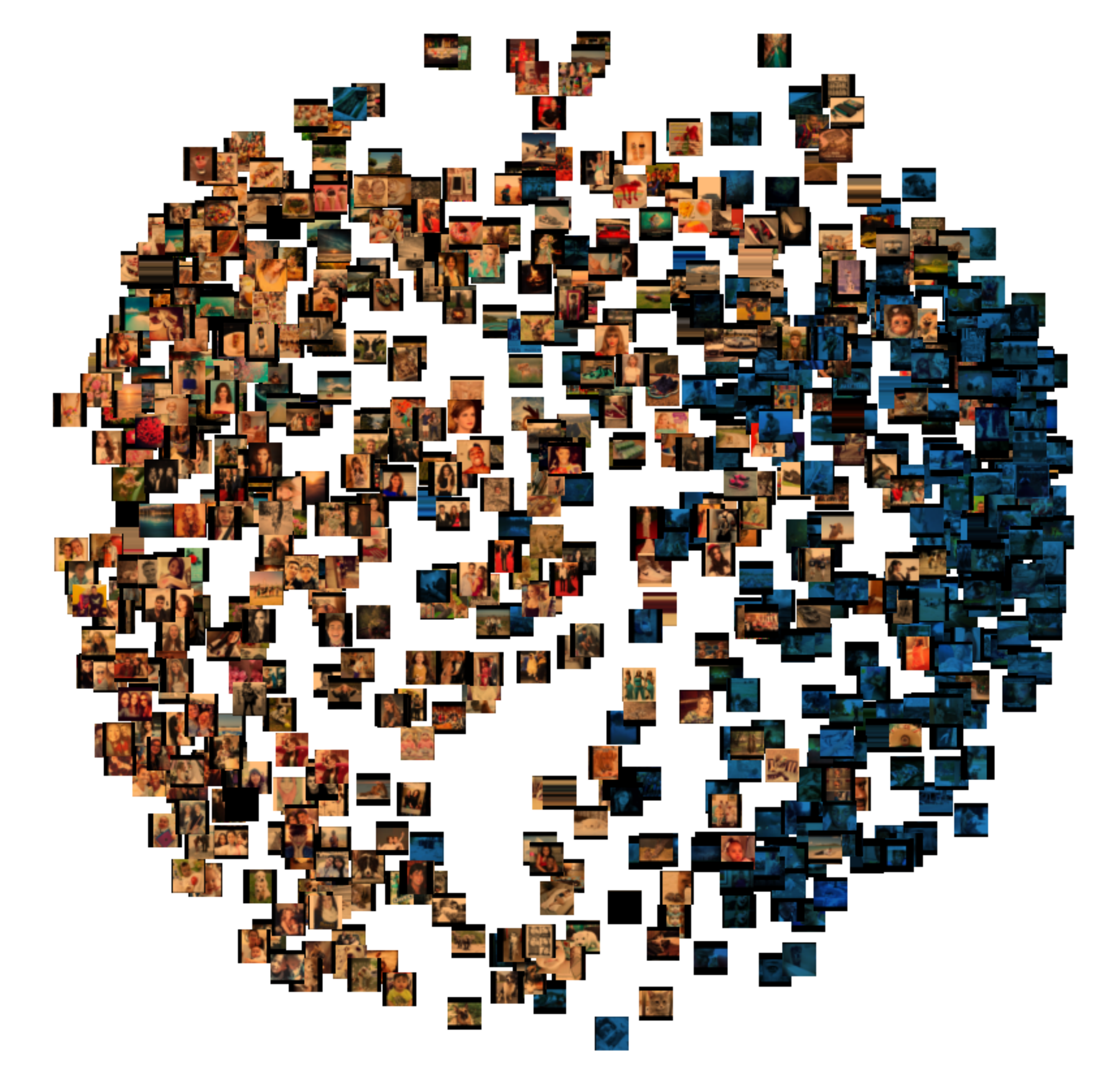}    
	\caption{Low dimensional representation using the first two principal components of the emoji embedding and the corresponding sentiment label (blue for negative and yellow for positive sentiment).}
\label{fig:emoji_sent_pca}
\vspace{-0.3cm}
\end{figure}
 
\begin{figure*}[!t]
	\centering
    \includegraphics[width=0.75\linewidth]{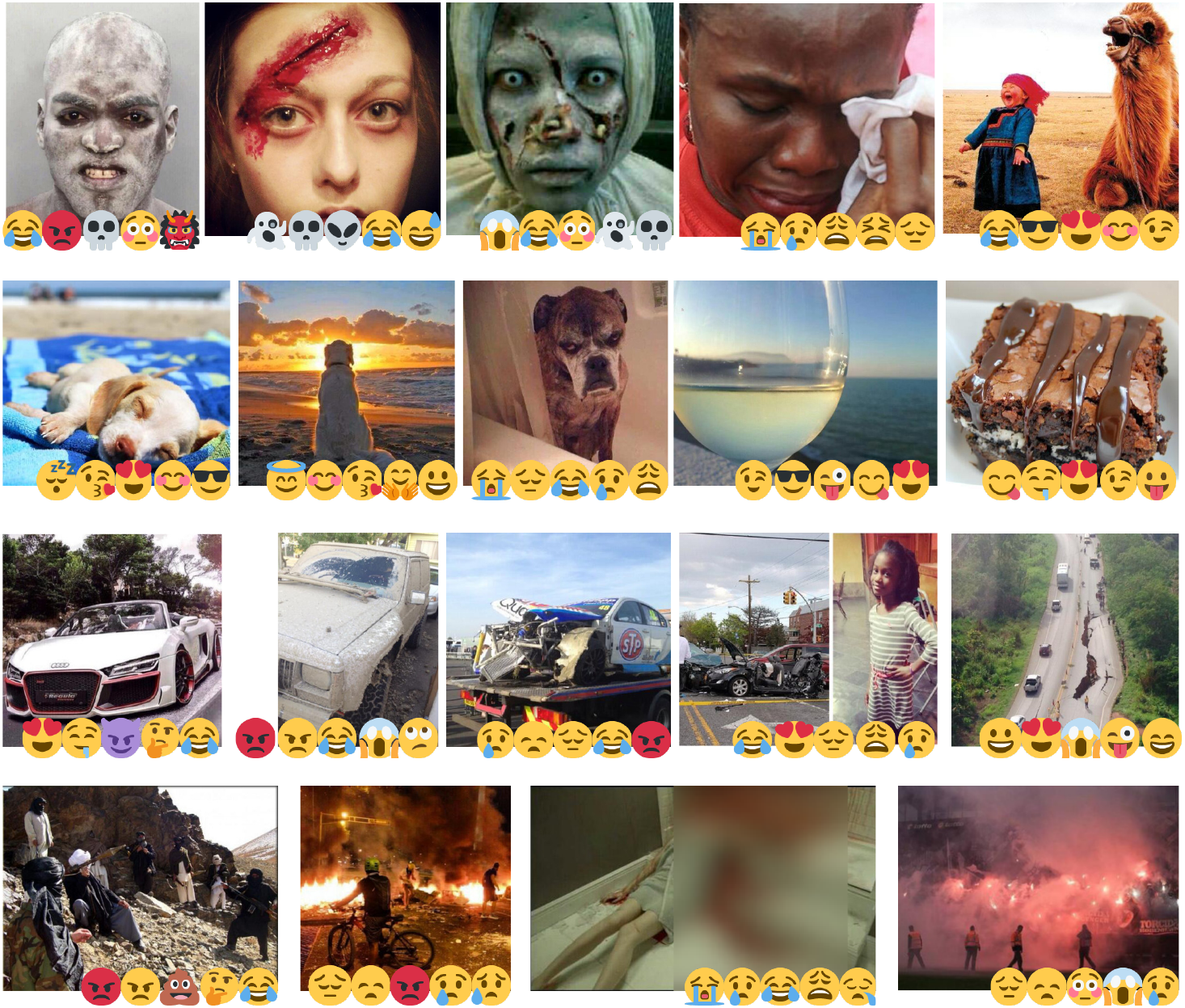}
	\caption{Qualitative results for the top 5 emojis predicted per image using our SmileyNet (ordered left to right).
	In contrast to a \textit{sentiment neutral} object representation, our model produces diverse output for objects of the same category depending on the emotion conveyed in the image, \eg see predictions on faces, dogs \& cars in $1^{st}$, $2^{nd}$ \& $ 3^{rd}$ rows. }
\label{fig:qual_res}
\vspace{-0.3cm}
\end{figure*}
 \paragraph{Qualitative results}
\figref{fig:qual_res} shows the top predictions of our SmileyNet for some test images from Twitter~\cite{you2015robust}.
Our model produces sensible predictions that capture the general sentiment in the image.
Unlike a model trained for object classification, SmileyNet output is not tailored to the object category but rather to the sentiment depicted by the object.
This can be best observed by checking the model output for similar objects, like the faces, the dogs and the cars images.
Our model predicts emojis of sentiment with opposite polarities when the input image is composed of sub-images (like the car accident and the child, $3^{rd}$ row) or when the main sentiment region is not in focus (like the image of the damaged road, $3^{rd}$ row).
This can be related to the holistic approach of the SmileyNet.
We hypothesize that an attention or region based processing might help in prioritizing the most influential image area for final predictions.
Finally, predictions on images similar to those in the $4^{th}$ row, suggest that SmileyNet might be helpful not only for sentiment analysis but also for novel applications such as detecting violence or abuse in images.

\subsection{Visual Sentiment}\label{eval:vis_sent}
\vspace{0.2cm}
\paragraph{Dataset}
We evaluate our model on the Twitter dataset~\cite{you2015robust}.
The dataset contains $1269$ images collected from Twitter and labeled manually by several annotators with positive and negative sentiment.
It has $3$ splits based on the degree of agreement among the annotators: ``5 agrees'', ``4 agrees'', and ``3 agrees''.
For example, \textit{4 agrees} split has images that at least 4 human annotators agreed upon their sentiment label.

\paragraph{Emojis \& sentiment}
We use our SmileyNet to embed all images of the ``5 agrees'' split in the emoji space without \textit{any} further training.
\figref{fig:emoji_sent_pca} shows the projection of these embeddings in 2D using the first $2$ dimensions of principle component analysis (PCA).
One can clearly see that samples of both positive and negative sentiments are well separated in this low dimensional space.
This indicates that our emoji embedding does indeed capture the visual sentiment exhibited in the image.
Furthermore, using the Spearman's rank-order correlation analysis, we analyze the relations between the individual emoji dimensions and the sentiment labels.
We find out that emojis with the highest correlation with the \textit{positive} sentiment are: (\emoji{015}, $0.62$), (\emoji{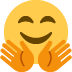},$0.62$), (\emoji{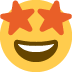}, $0.58$), (\emoji{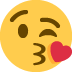}, $0.56$) and (\emoji{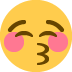}, $0.53$), whereas emojis with the highest correlation to \textit{negative} sentiment are: (\emoji{045}, $0.67$), (\emoji{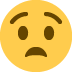},$0.66$), (\emoji{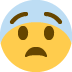}, $0.65$), (\emoji{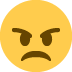}, $0.64$) and (\emoji{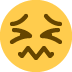}, $0.64$).

\begin{table}
\center
\scalebox{0.85}{
\begin{tabular}{ l c c c}
\toprule
  							& \multicolumn{3}{c}{Twitter Visual Sentiment~\cite{you2015robust}}\\
Model 						& 	3 agrees & 	4 agrees & 	5 agrees \\
\midrule 			
ObjectNet	 				& 	74.0		&		79.0	&	82.1		\\
SmileyNet (ours)			& 	\textbf{76.5} 	& 	\textbf{80.0} 	& 	\textbf{84.7}   \\
\bottomrule
\end{tabular}
}
\caption{$1$-Nearest neighbor sentiment prediction accuracy.}
\label{tbl:sent_pred_nn}
\vspace{-0.4cm}
\end{table} \paragraph{Emojis \& objects}
To evaluate the quality of the embedding quantitatively, we use $1$ nearest neighbor classification and do 5-fold cross validation over the sentiment dataset for each of the $3$ splits.
We compare our emoji-embedding to an embedding produced by a model with the same base architecture (\ie ResNet50) but trained over the ImageNet dataset (ObjectNet).
As expected, our SmileyNet produces better embeddings for sentiment analysis than ObjectNet and outperforms it on all three splits (see \tblref{tbl:sent_pred_nn}), while SmileyNet's embedding is $10$ times smaller compared to that of ObjectNet.

\paragraph{Transfer learning}
Alternatively, we can adopt a transfer learning scheme and finetune our model on the target set to see how well our model can adapt to the target data distribution from a few samples.
We realize $t(\cdot)$ as a fully connected layer (\secref{sec:model}) and use 5-fold cross validation to finetune and test our model as in \cite{you2015robust}.

\begin{table}
\center
\scalebox{0.80}{
\begin{tabular}{ l c c c}
\toprule
  							& \multicolumn{3}{c}{Twitter Visual Sentiment~\cite{you2015robust}} \\
Model 										& 	3 agrees & 	4 agrees & 	5 agrees \\
\midrule
PAEF~\cite{zhao2014exploring}				&	67.92	&	69.61 	& 	72.90 \\ 
SentiBank~\cite{borth2013large} 			& 	66.63 	& 	68.28 	& 	71.32 \\ 
DeepSentiBank~\cite{chen2014deepsentibank}	& 	71.25 	& 	70.15 	& 	76.35 \\ 
PCNN~\cite{you2015robust}					& 	76.36 	& 	76.52 	& 	82.54 \\ 
Campos \etal~\cite{campos2017pixels}						&	74.90	& 	78.70	&	83.00 \\ 
AR+Concat(K=1)~\cite{yang2018visual}		&	77.79 	& 	83.25 	& 	86.10 \\ 
AR+Concat(K=8)~\cite{yang2018visual}		&	81.06 	& 	\textbf{85.10} 	& 	88.65 \\ 
\midrule
ObjectNet 									& 	78.28	&	82.73	&	87.67		\\
SmileyNet (ours)							& 	\textbf{82.69} 	& 	\underline{84.87} 	& 	\textbf{89.16}   \\
\bottomrule
\end{tabular}
}
\caption{State-of-the-art comparison of SimleyNet for visual sentiment prediction.}
\label{tbl:sent_pred_ft}
\vspace{-0.2cm}
\end{table} \tblref{tbl:sent_pred_ft} compares the accuracy of our model to state-of-the-art (SOTA) models.
Our SimleyNet outperforms the SentiBank models \cite{borth2013large,chen2014deepsentibank} which embed images in Adjective-Noun pairs (ANP) space that is learned as well from social media data.
This indicate that emojis are better in capturing sentiment than text-based cues.
We speculate emoji labeling has the advantage of being universal, finite, and offers an unambiguous one-to-one mapping between label and emotion, whereas words carry rich connotations that may make the design of an effective lexicon mapping words to emotions more difficult.
Moreover, our SmileyNet outperforms the advanced AR model~\cite{yang2018visual} that employs a customized approach with attention mechanisms when using a single model ($K=1$), like ours, and even when using an ensemble of $K=8$ models.
This is significant given that our model leverages off-the-shelf neural architecture and trained using noisy social media data.
This further demonstrates the effectiveness of the learned embedding.
We hypothesize that our model can be improved even further by employing an ensemble of models like in~\cite{yang2018visual} or customized attention modules such as~\cite{Fan2017}.

\begin{table}
\center
\scalebox{0.85}{
\begin{tabular}{ l c c c}
\toprule
  							& \multicolumn{3}{c}{Twitter Visual Sentiment~\cite{you2015robust}}\\
Model 						& 	3 agrees & 	4 agrees & 	5 agrees \\
\midrule
SmileyNet - Con.			&   73.4 	 &  76.0 	 &  80.0 \\
SmileyNet - Bin.			&   74.2 	 &  77.1 	 &  81.2 \\
\bottomrule
\end{tabular}
}
\caption{Zero-shot visual sentiment prediction accuracy.}
\label{tbl:sent_pred_zsl}
\vspace{-0.4cm}
\end{table}
 \paragraph{Zero-shot visual sentiment prediction}
Unlike other representations, our embedding is interpretable and each dimension can be easily related to a certain sentiment class.
That is we can construct a sentiment classifier without using any training images, \ie zero-shot learning (ZSL) \cite{Lampert2009,Al-Halah2016}.
To our knowledge, ours is the \emph{first} work to attempt ZSL for visual sentiment.
We ask $4$ annotators to label each of the emojis in our representation with a positive or negative sentiment based solely on the emoji's visual depiction.
Then we use the average annotation as a mapping $t(\cdot)$ that will ensemble the emoji's prediction scores to estimate whether an image $x$ has a positive or a negative sentiment.
\tblref{tbl:sent_pred_zsl} shows the performance of our model in ZSL setting.
Interestingly, while using \emph{no training images} at all our model is still capable of producing reliable sentiment prediction that is competitive with many of the SOTA models in \tblref{tbl:sent_pred_ft}.
We also see that using equal weighting to each emoji (the binary version ``Bin.'') lead to higher accuracy in comparison to using the average annotation to weight the emoji's prediction in the ensemble (the continuous model ``Con.'').

\subsection{Fine-grained Emotions}
\vspace{0.2cm}
\paragraph{Dataset}
Finally, we evaluate our model for fine-grained emotion classification on the Flickr\&Instagram dataset~\cite{you2016building}.
The dataset contains $23,308$ images queried from Flickr and Instagram and labeled by Amazon Mechanical Turk with $8$ emotion classes: \textit{amusement}, \textit{anger}, \textit{awe}, \textit{contentment}, \textit{disgust}, \textit{excitement}, \textit{fear} and \textit{sadness}.

\begin{table}
\center
\scalebox{0.75}{
\begin{tabular}{ l c c c c c c c}
\toprule
Emotion 	& 	\multicolumn{7}{c}{Most Correlated Emojis} \\
\midrule
amusement 	& 	\emoji{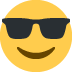}	& \emoji{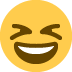}	& \emoji{003}	&	\emoji{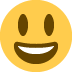} &	\emoji{029} &	\emoji{000} & \emoji{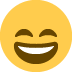}		\\
& 	0.31	& 0.30	& 0.29 	&	0.27 &	0.27 &	0.26 & 0.26 \\
\hline
anger		& 	\emoji{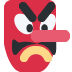} 	& \emoji{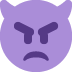} 	& \emoji{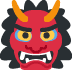} &  \emoji{089} & \emoji{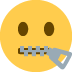} & \emoji{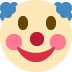}  & \emoji{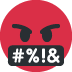}	   \\
& 	0.18 	& 0.18 	& 0.17 &  0.17 & 0.16 & 0.16  & 0.15	   \\
\hline
awe 		& 	\emoji{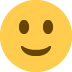} 	& \emoji{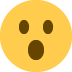} & \emoji{009} 	& 	 \emoji{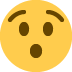} & \emoji{000} & \emoji{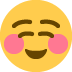} & \emoji{001} \\
& 	0.28 	& 0.24 & 0.24 	& 	 0.23 & 0.23 & 0.22 & 0.21 \\
\hline
contentment & 	 \emoji{016}	& \emoji{019} & \emoji{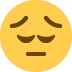} & \emoji{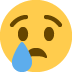} & \emoji{018} & \emoji{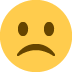} & \emoji{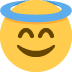}   \\
& 	 0.27	& 0.26 & 0.24 & 0.24 & 0.23 & 0.23 & 0.22   \\
\hline
disgust 	& \emoji{077} & \emoji{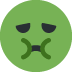} & \emoji{075} & \emoji{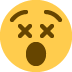} & \emoji{039} & \emoji{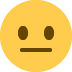} & \emoji{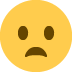} \\
& 0.29 & 0.28 & 0.23 & 0.20 & 0.20 & 0.18 & 0.17 \\
\hline
excitement 	& \emoji{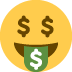} & \emoji{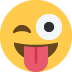} & \emoji{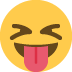} & \emoji{029} & \emoji{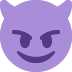} & \emoji{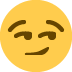} & \emoji{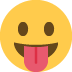} \\
& 0.22 & 0.20 & 0.20 & 0.19 & 0.17 & 0.17 & 0.17 \\
\hline
fear 		& \emoji{089} & \emoji{090} & \emoji{084} & \emoji{083} & \emoji{085} & \emoji{088} & \emoji{086}\\
& 0.21 & 0.18 & 0.17 & 0.17 & 0.16 & 0.16 & 0.15\\
\hline
sadness 	& \emoji{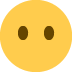} & \emoji{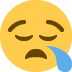} & \emoji{033} & \emoji{089} & \emoji{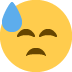} & \emoji{039} & \emoji{076}\\
& 0.26 & 0.25 & 0.24 & 0.23 & 0.22 & 0.21 & 0.21\\
\bottomrule
\end{tabular}
}
\caption{Top correlated Emojis with each emotion class.}
\label{tbl:emotion_corr}
\end{table}

 \paragraph{Emojis \& emotions}
We analyze first the correlations between emojis and emotion classes.
\tblref{tbl:emotion_corr} ranks the most correlated emojis per emotion class.
Interestingly, many of the top ranked emojis correspond to our intuition of the emotion depicted by the emoji's image itself.
However, the ranking also reveals some unexpected correlations like \emoji{086} with \textit{anger} and \textit{fear}, \emoji{060} with \textit{disgust}, \emoji{073} with \textit{anger}, and \emoji{071} with \textit{sadness}.
Some of these come form cultural context (like \emoji{086}), while others we expect from common confusion of similarly looking emojis (like the sleepy face \emoji{071} and crying face \emoji{042}).

\begin{table}
\center
\scalebox{0.80}{
\begin{tabular}{ l c c c}
\toprule
Model 							& 	Multi-Class 	& 	Sentiment \\
&Emotions& \\
\midrule
You \etal~\cite{you2016building} 			& 	48.30					& - 	\\
DeepSentiBank~\cite{chen2014deepsentibank}				
								& 	-						& 61.54 \\
PCNN~\cite{you2015robust}		& 	-						& 75.34 \\
AR+Concat(K=1)~\cite{yang2018visual}		
								&	- 						& 84.83	\\
AR+Concat(K=8)~\cite{yang2018visual}		
								&	- 						& 86.35	\\
\midrule
ObjectNet 						& 	54.42					& 83.81	\\
SmileyNet (ours)				& 	\textbf{55.81} 			& \textbf{87.01}    \\
\bottomrule
\end{tabular}
}
\caption{Fine-grained emotion classification accuracy on the Flickr\&Instagram dataset~\cite{you2016building}.}
\label{tbl:emotion_pred_ft}
\vspace{-0.2cm}
\end{table} 
\begin{figure}[!t]
	\centering
    \includegraphics[width=0.75\linewidth]{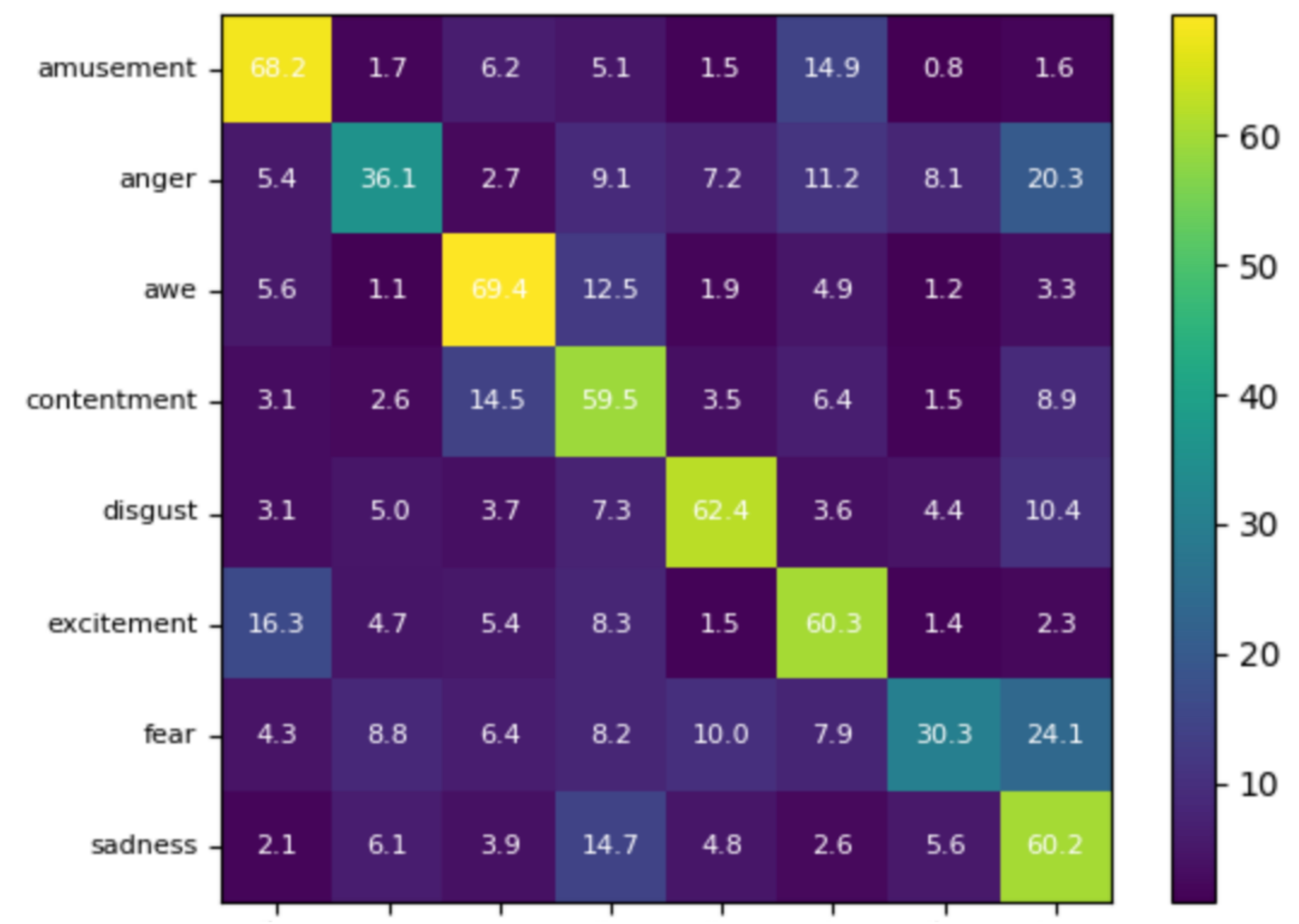}    
	\caption{Confusion matrix of our SmileyNet predictions of the 8 emotion classes in Flickr\&Instagram dataset.}
\label{fig:emotion_conf_mat}
\end{figure}
 \paragraph{Transfer learning}
\tblref{tbl:emotion_pred_ft} shows the performance of our SmileyNet in predicting the $8$ emotion classes in a transfer learning setting.
Similar to the previous section, we compare our model to ObjectNet which has been trained previously on the ImageNet dataset as it is commonly the case in literature.
As hypothesized previously, SmileyNet is more suitable for fine-grained emotion prediction and outperforms a similar model transferred from an object classification task (\ie ObjectNet).
Moreover, our model outperforms SOTA in this task as well and shows that our compact embedding is highly effective for fine-grained emotion prediction.
\figref{fig:emotion_conf_mat} gives us a deeper insight on the performance of each of the emotion classes.
Most of the emotions are predicted with equal accuracy except for \textit{anger} and \textit{fear} which show high confusion with the \textit{sadness}.
Finally, similar to \cite{yang2018visual}, we map the $8$ emotion classes to positive and negative sentiment and report classification accuracy.
Our model outperforms SOTA for this derivative task too, in accordance to our previous results from \secref{eval:vis_sent}.

\subsection{Emoji's Emotional Fingerprint}

\begin{figure}[!t]
\centering
\begin{subfigure}[!b]{0.3\linewidth}
    \centering
    \includegraphics[width=\linewidth]{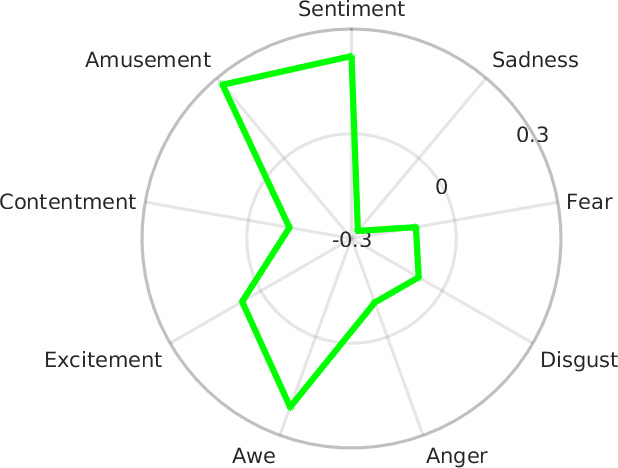}    
    \caption{\emoji{001}}
\end{subfigure}
\begin{subfigure}[!b]{0.3\linewidth}
    \centering
    \includegraphics[width=\linewidth]{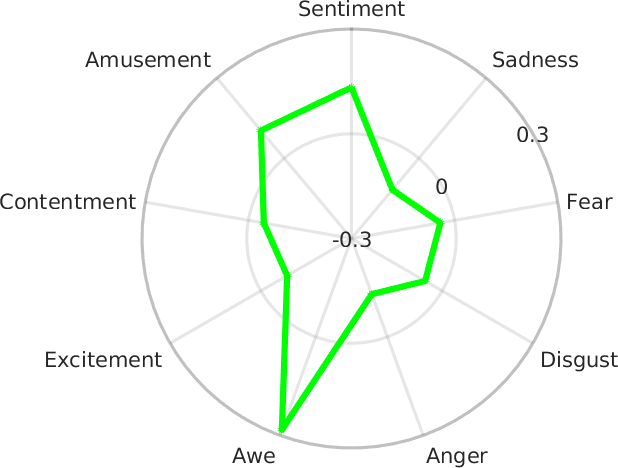}    
    \caption{\emoji{011}}
\end{subfigure}
\begin{subfigure}[!b]{0.3\linewidth}
    \centering
    \includegraphics[width=\linewidth]{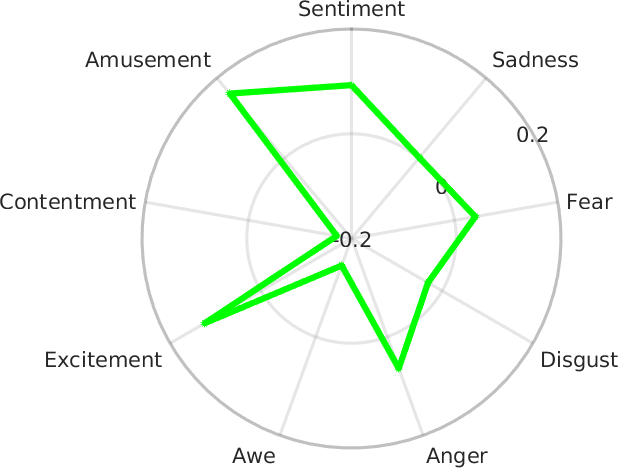}    
    \caption{\emoji{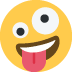}}
\end{subfigure}\\
\begin{subfigure}[!b]{0.3\linewidth}
    \centering
    \includegraphics[width=\linewidth]{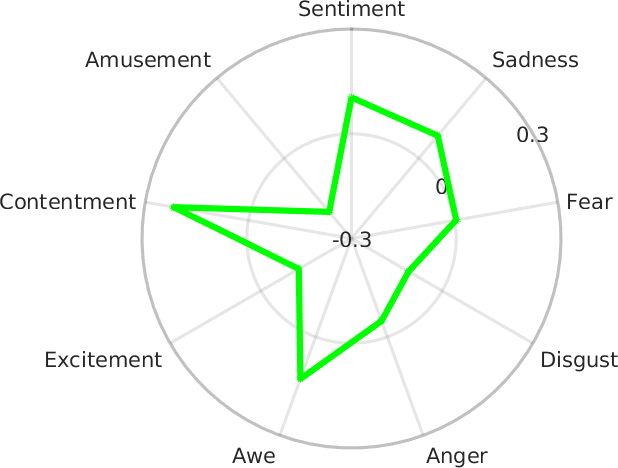}    
    \caption{\emoji{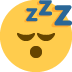}}
\end{subfigure}
\begin{subfigure}[!b]{0.3\linewidth}
    \centering
    \includegraphics[width=\linewidth]{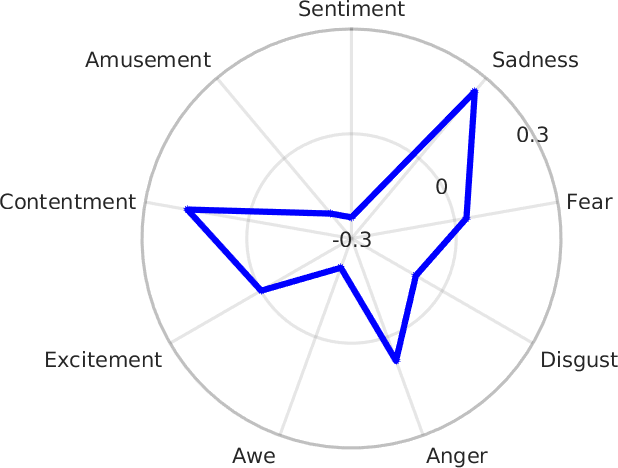}    
    \caption{\emoji{071}}
\end{subfigure}
\begin{subfigure}[!b]{0.3\linewidth}
    \centering
    \includegraphics[width=\linewidth]{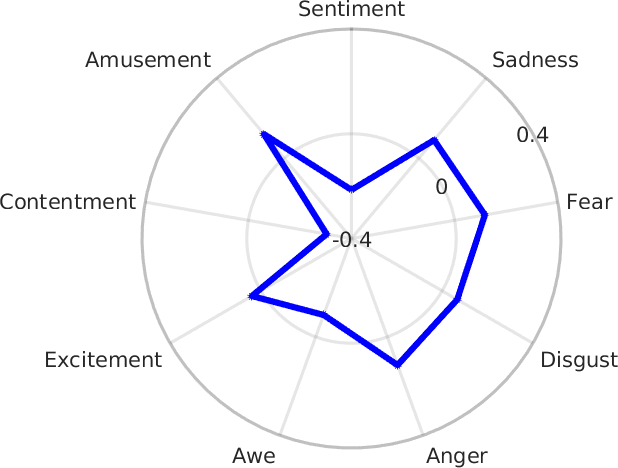}    
    \caption{\emoji{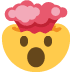}}
\end{subfigure}
\caption{Emoji's emotional fingerprint. Our model reveals a unique emotional response for each emoji. Fingerprints with general positive or negative sentiment are colored with green and blue respectively.}
\label{fig:emojis_emotions}
\vspace{-0.3cm}
\end{figure}
 
\begin{figure}[!t]
	\centering
	\setlength{\fboxsep}{0pt}
	\setlength{\fboxrule}{0.5pt}
	\fbox{\includegraphics[width=0.6\linewidth]{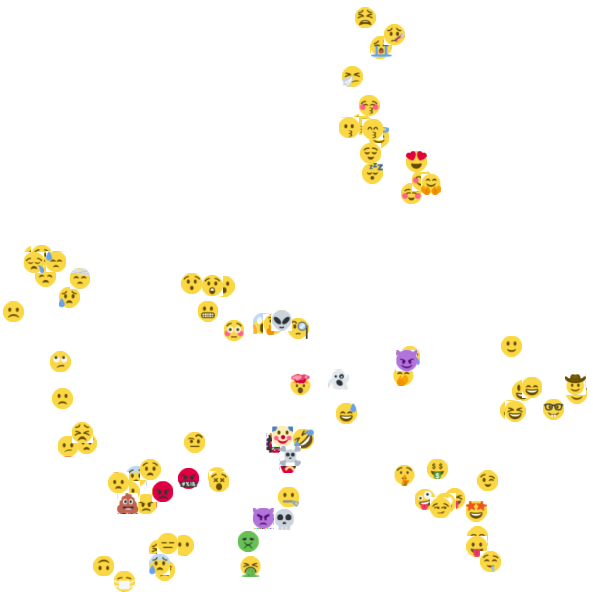}}
	\caption{Low dimensional embedding of the emojis using t-SNE and based on their emotional fingerprint.}
\label{fig:emojis_vefp_tsne}
\vspace{-0.3cm}
\end{figure}
 
Given our previous analysis, we notice that each emoji in our representation has a unique signature in the emotional space.
\figref{fig:emojis_emotions} shows a sample of $6$ emojis and their corresponding emotional fingerprint (EEF).
We see that even emoji that have similar portrayal such as \emoji{001} \& \emoji{011} or similar semantics like the sleepy \emoji{071} \& sleeping \emoji{069} face have different emotional response both in intensity and bias towards certain type of emotions.
Furthermore, projecting the emojis in 2D space based on their emotional fingerprints reveals further interesting findings (\figref{fig:emojis_vefp_tsne}).
For example, [\emoji{006}\emoji{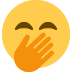}\emoji{082}] has similar EEF, the EEF of~\emoji{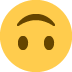} is closer to \emoji{077}\emoji{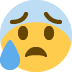} than to \emoji{011}, and \emoji{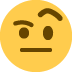} shows bias towards anger, disgust and fear in its EEF similar to \emoji{046}\emoji{072}\emoji{066}.
We believe this novel representation can be of great interest for further research in behavioral studies in social media and deeper understanding of the emoji modality and its usage.

\section{Conclusion}
We propose to circumvent current limitations of small visual sentiment analysis datasets by learning a compact image embedding from readily available data in social media.
Unlike the common object-based embedding, the proposed embedding is well aligned with the visual sentiment label space and generalizes better in transfer learning settings.
Furthermore, our embedding can be efficiently learned from noisy data in social media by leveraging the intricate relation between emojis and images.
To that end, we build a novel dataset, the Visual Smiley Dataset, which we use to learn an emoji-based image embedding.
The evaluation on sentiment and emotion recognition shows that our low-dimensional embedding consistently outperforms the commonly used object-based embedding and the more elaborate and customized SOTA models.
Furthermore, due to its interpretability we demonstrate that our embedding can be used for sentiment analysis without any further training in a zero-shot learning setting.
Finally, initial results show that our embedding can aid novel applications for which inferring emotion from visual data is relevant, \eg visual abuse and violence detection.
We expect this work findings to be of interest not only for computer vision and visual sentiment analysis communities but also for social media studies and emoji modality understanding.

\clearpage
\balance
{\small
\bibliographystyle{ieee}
\bibliography{mybib}

\begin{thebibliography}{10}\itemsep=-1pt

\bibitem{Al-Halah2016}
Z.~Al-Halah, M.~Tapaswi, and R.~Stiefelhagen.
\newblock {Recovering the missing link: Predicting class-attribute associations
  for unsupervised zero-shot learning}.
\newblock In {\em CVPR}, 2016.

\bibitem{alameda2016recognizing}
X.~Alameda-Pineda, E.~Ricci, Y.~Yan, and N.~Sebe.
\newblock Recognizing emotions from abstract paintings using non-linear matrix
  completion.
\newblock In {\em CVPR}, 2016.

\bibitem{baldwin2013noisy}
T.~Baldwin, P.~Cook, M.~Lui, A.~MacKinlay, and L.~Wang.
\newblock How noisy social media text, how diffrnt social media sources?
\newblock In {\em Proceedings of the Sixth International Joint Conference on
  Natural Language Processing}, 2013.

\bibitem{barbieri2016cosmopolitan}
F.~Barbieri, G.~Kruszewski, F.~Ronzano, and H.~Saggion.
\newblock How cosmopolitan are emojis?: Exploring emojis usage and meaning over
  different languages with distributional semantics.
\newblock In {\em ACM MM}, 2016.

\bibitem{barbieri2016does}
F.~Barbieri, F.~Ronzano, and H.~Saggion.
\newblock What does this emoji mean? a vector space skip-gram model for twitter
  emojis.
\newblock In {\em LREC}, 2016.

\bibitem{borth2013large}
D.~Borth, R.~Ji, T.~Chen, T.~Breuel, and S.-F. Chang.
\newblock Large-scale visual sentiment ontology and detectors using adjective
  noun pairs.
\newblock In {\em ACM MM}, 2013.

\bibitem{campos2017pixels}
V.~Campos, B.~Jou, and X.~Giro-i Nieto.
\newblock From pixels to sentiment: Fine-tuning cnns for visual sentiment
  prediction.
\newblock {\em Image and Vision Computing}, 65:15--22, 2017.

\bibitem{cappallo2018new}
S.~Cappallo, S.~Svetlichnaya, P.~Garrigues, T.~Mensink, and C.~G. Snoek.
\newblock New modality: Emoji challenges in prediction, anticipation, and
  retrieval.
\newblock {\em IEEE Transactions on Multimedia}, 21(2):402--415, 2018.

\bibitem{chen2014deepsentibank}
T.~Chen, D.~Borth, T.~Darrell, and S.-F. Chang.
\newblock Deepsentibank: Visual sentiment concept classification with deep
  convolutional neural networks.
\newblock {\em arXiv preprint arXiv:1410.8586}, 2014.

\bibitem{eisner2016emoji2vec}
B.~Eisner, T.~Rockt{\"a}schel, I.~Augenstein, M.~Bo{\v{s}}njak, and S.~Riedel.
\newblock emoji2vec: Learning emoji representations from their description.
\newblock {\em arXiv preprint arXiv:1609.08359}, 2016.

\bibitem{el2017face2emoji}
A.~El~Ali, T.~Wallbaum, M.~Wasmann, W.~Heuten, and S.~C. Boll.
\newblock Face2emoji: Using facial emotional expressions to filter emojis.
\newblock In {\em Proceedings of the 2017 CHI Conference Extended Abstracts on
  Human Factors in Computing Systems}. ACM, 2017.

\bibitem{emoji_stats}
Emojipedia.
\newblock {Emoji statistics}.
\newblock \url{https://worldemojiday.com/statistics}, 2018.
\newblock [Online; accessed 25-Feb-2019].

\bibitem{Fan2017}
S.~Fan, M.~Jiang, Z.~Shen, B.~L. Koenig, M.~S. Kankanhalli, and Q.~Zhao.
\newblock {The Role of Visual Attention in Sentiment Prediction}.
\newblock In {\em ACM MM}, 2017.

\bibitem{felbo2017using}
B.~Felbo, A.~Mislove, A.~S{\o}gaard, I.~Rahwan, and S.~Lehmann.
\newblock Using millions of emoji occurrences to learn any-domain
  representations for detecting sentiment, emotion and sarcasm.
\newblock {\em arXiv preprint arXiv:1708.00524}, 2017.

\bibitem{glorot2011domain}
X.~Glorot, A.~Bordes, and Y.~Bengio.
\newblock Domain adaptation for large-scale sentiment classification: A deep
  learning approach.
\newblock In {\em ICML}, 2011.

\bibitem{go2009twitter}
A.~Go, R.~Bhayani, and L.~Huang.
\newblock Twitter sentiment classification using distant supervision.
\newblock {\em CS224N Project Report, Stanford}, 1(12), 2009.

\bibitem{guthier2017language}
B.~Guthier, K.~Ho, and A.~El~Saddik.
\newblock Language-independent data set annotation for machine learning-based
  sentiment analysis.
\newblock In {\em IEEE International Conference on Systems, Man, and
  Cybernetics (SMC)}, 2017.

\bibitem{he2016deep}
K.~He, X.~Zhang, S.~Ren, and J.~Sun.
\newblock Deep residual learning for image recognition.
\newblock In {\em CVPR}, 2016.

\bibitem{kelly2015characterising}
R.~Kelly and L.~Watts.
\newblock Characterising the inventive appropriation of emoji as relationally
  meaningful in mediated close personal relationships.
\newblock {\em Experiences of Technology Appropriation: Unanticipated Users,
  Usage, Circumstances, and Design}, 2015.

\bibitem{kim2018building}
H.-R. Kim, Y.-S. Kim, S.~J. Kim, and I.-K. Lee.
\newblock Building emotional machines: Recognizing image emotions through deep
  neural networks.
\newblock {\em IEEE Transactions on Multimedia}, 2018.

\bibitem{Kingma2015}
D.~P. Kingma and J.~L. Ba.
\newblock {ADAM: A Method for Stochastic Optimization}.
\newblock In {\em ICLR}, 2015.

\bibitem{Lampert2009}
C.~H. Lampert, H.~Nickisch, and S.~Harmeling.
\newblock {Learning to detect unseen object classes by between-class attribute
  transfer}.
\newblock In {\em CVPR}, 2009.

\bibitem{lang2005international}
P.~J. Lang.
\newblock International affective picture system (iaps): Affective ratings of
  pictures and instruction manual.
\newblock {\em Technical report}, 2005.

\bibitem{lebduska2014emoji}
L.~Lebduska.
\newblock Emoji, emoji, what for art thou?
\newblock {\em Harlot: A Revealing Look at the Arts of Persuasion}, 1(12),
  2014.

\bibitem{liu2012survey}
B.~Liu and L.~Zhang.
\newblock A survey of opinion mining and sentiment analysis.
\newblock In {\em Mining text data}, pages 415--463. Springer, 2012.

\bibitem{ljubevsic2016global}
N.~Ljube{\v{s}}i{\'c} and D.~Fi{\v{s}}er.
\newblock A global analysis of emoji usage.
\newblock In {\em Proceedings of the 10th Web as Corpus Workshop}, 2016.

\bibitem{lu2012shape}
X.~Lu, P.~Suryanarayan, R.~B. Adams~Jr, J.~Li, M.~G. Newman, and J.~Z. Wang.
\newblock On shape and the computability of emotions.
\newblock In {\em ACM MM}, 2012.

\bibitem{machajdik2010affective}
J.~Machajdik and A.~Hanbury.
\newblock Affective image classification using features inspired by psychology
  and art theory.
\newblock In {\em ACM MM}, 2010.

\bibitem{medhat2014sentiment}
W.~Medhat, A.~Hassan, and H.~Korashy.
\newblock Sentiment analysis algorithms and applications: A survey.
\newblock {\em Ain Shams Engineering Journal}, 5(4):1093--1113, 2014.

\bibitem{mikels2005emotional}
J.~A. Mikels, B.~L. Fredrickson, G.~R. Larkin, C.~M. Lindberg, S.~J. Maglio,
  and P.~A. Reuter-Lorenz.
\newblock Emotional category data on images from the international affective
  picture system.
\newblock {\em Behavior research methods}, 37(4):626--630, 2005.

\bibitem{miller2017understanding}
H.~J. Miller, D.~Kluver, J.~Thebault-Spieker, L.~G. Terveen, and B.~J. Hecht.
\newblock Understanding emoji ambiguity in context: The role of text in
  emoji-related miscommunication.
\newblock In {\em ICWSM}, 2017.

\bibitem{mohammad2018semeval}
S.~Mohammad, F.~Bravo-Marquez, M.~Salameh, and S.~Kiritchenko.
\newblock Semeval-2018 task 1: Affect in tweets.
\newblock In {\em Proceedings of The 12th International Workshop on Semantic
  Evaluation}, 2018.

\bibitem{ng2015deep}
H.-W. Ng, V.~D. Nguyen, V.~Vonikakis, and S.~Winkler.
\newblock Deep learning for emotion recognition on small datasets using
  transfer learning.
\newblock In {\em Proceedings of the International Conference on Multimodal
  Interaction}. ACM, 2015.

\bibitem{novak2015sentiment}
P.~K. Novak, J.~Smailovi{\'c}, B.~Sluban, and I.~Mozeti{\v{c}}.
\newblock Sentiment of emojis.
\newblock {\em PloS one}, 10(12):e0144296, 2015.

\bibitem{peng2015mixed}
K.-C. Peng, T.~Chen, A.~Sadovnik, and A.~C. Gallagher.
\newblock A mixed bag of emotions: Model, predict, and transfer emotion
  distributions.
\newblock In {\em CVPR}, 2015.

\bibitem{rathan2018consumer}
M.~Rathan, V.~R. Hulipalled, K.~Venugopal, and L.~Patnaik.
\newblock Consumer insight mining: aspect based twitter opinion mining of
  mobile phone reviews.
\newblock {\em Applied Soft Computing}, 68:765--773, 2018.

\bibitem{read2005using}
J.~Read.
\newblock Using emoticons to reduce dependency in machine learning techniques
  for sentiment classification.
\newblock In {\em Proceedings of the ACL student research workshop}, 2005.

\bibitem{russakovsky2015imagenet}
O.~Russakovsky, J.~Deng, H.~Su, J.~Krause, S.~Satheesh, S.~Ma, Z.~Huang,
  A.~Karpathy, A.~Khosla, M.~Bernstein, et~al.
\newblock Imagenet large scale visual recognition challenge.
\newblock {\em International Journal of Computer Vision}, 115(3):211--252,
  2015.

\bibitem{santhanam2018stand}
S.~Santhanam, V.~Srinivasan, S.~Glass, and S.~Shaikh.
\newblock I stand with you: Using emojis to study solidarity in crisis events.
\newblock In {\em Proceedings of the 1st International Workshop on Emoji
  Understanding and Applications in Social Media}, 2018.

\bibitem{tang2014coooolll}
D.~Tang, F.~Wei, B.~Qin, T.~Liu, and M.~Zhou.
\newblock Coooolll: A deep learning system for twitter sentiment
  classification.
\newblock In {\em Proceedings of the 8th international workshop on semantic
  evaluation (SemEval 2014)}, 2014.

\bibitem{emojiv12}
Unicode.
\newblock Emoji list v12.0.
\newblock \url{https://unicode.org/emoji/charts/emoji-counts.html}, 2019.
\newblock [Online; accessed 25-Feb-2019].

\bibitem{yang2018visual}
J.~Yang, D.~She, M.~Sun, M.-M. Cheng, P.~Rosin, and L.~Wang.
\newblock Visual sentiment prediction based on automatic discovery of affective
  regions.
\newblock {\em IEEE Transactions on Multimedia}, 2018.

\bibitem{you2015robust}
Q.~You, J.~Luo, H.~Jin, and J.~Yang.
\newblock Robust image sentiment analysis using progressively trained and
  domain transferred deep networks.
\newblock In {\em AAAI}, 2015.

\bibitem{you2016building}
Q.~You, J.~Luo, H.~Jin, and J.~Yang.
\newblock Building a large scale dataset for image emotion recognition: The
  fine print and the benchmark.
\newblock In {\em AAAI}, 2016.

\bibitem{zhao2014exploring}
S.~Zhao, Y.~Gao, X.~Jiang, H.~Yao, T.-S. Chua, and X.~Sun.
\newblock Exploring principles-of-art features for image emotion recognition.
\newblock In {\em ACM MM}, 2014.

\end{thebibliography}
}
\clearpage
\section{Appendix}

\begin{figure*}[b]
\centering
\includegraphics[width=.82\textwidth]{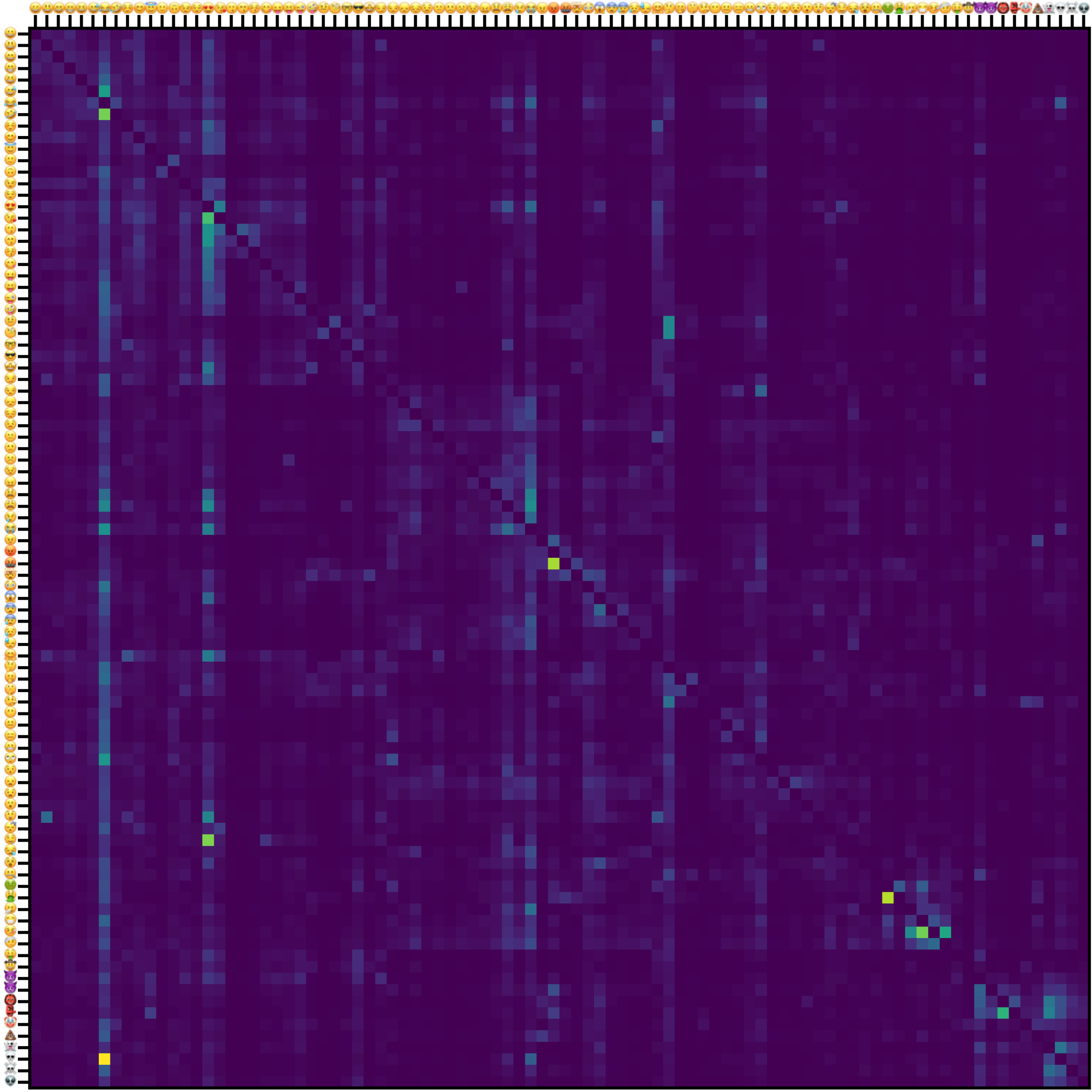}    
\caption{
	Label correlations in the proposed Visual Smiley Dataset. 
	Notice the high correlation of the popular \emph{face with tears of joy} and \emph{smiling face with heart eyes} with the majority of emojis (the two column in the first quarter of the matrix).
	Also the correlation matrix reveals semantically related groups suchs as \emph{ghost}, \emph{skull} and \emph{skull \& crossbones} and others (see the square structures in the lower right corner).
	Figure best viewed in color with zoom-in.}
\label{fig:emoji_corr}
\end{figure*}
 
\end{document}